# The Authority Resolution Framework

## A Five-Domain Ontology for Governing Who and What Decides, at Scale

**PARVIZ SHARIFF**
*Independent Researcher*, parviz.shariff1@gmail.com, parviz@portendlabs.com

## Abstract

I introduce the Authority Resolution Framework (ARF), a five-domain ontology for governing who and what holds the right to decide, at enterprise scale. Agentic AI is becoming the dominant axis of Enterprise IT spending, with industry forecasts projecting it will exceed a quarter of worldwide IT spend within a few years, even as a 2025 Forbes Research survey found fewer than one percent of executives report significant AI return on investment and Gartner has named unmanaged agentic AI proliferation, outpacing governance, its top 2026 cybersecurity trend. I argue this gap is a category-level allocation and governance problem, not a series of isolated technical failures, and that it cannot be solved by treating it as the Chief Information Officer (CIO) or Chief Digital Information Officer's (CDIO) responsibility alone.

Enterprises maintain at least five distinct, loosely coupled ontological representations of themselves:

1. A **Social** Structure of Roles and Informal Influence.
2. A **Business** Vocabulary of Domain terms.
3. A Codified Layer of Standardized **Processes**.
4. A **Machine**-readable Layer of Executable Code and Permissions.
5. An External **Real-World** Context the Enterprise does not control.

Decision rights and authority are conventionally treated as a property residing within the social or policy domain alone. I argue this is a structural error with increasing practical consequence as enterprises deploy agentic AI systems capable of autonomous action; an authority relation that is not explicitly resolved across all five domains simultaneously cannot be reliably or safely delegated to a non-human actor.

I propose the **Authority Relation** (AR) as a formal cross-domain primitive; a six-element tuple binding an actor, action, object, bounded context, justification chain, and a novel calibration term, the **DNA-Coefficient**, which quantifies the divergence between an organization's documented authority structure and its lived, practiced one. I show that the same cross-domain Object-resolution requirement underpinning the AR primitive yields semantic interoperability and consistent state-change propagation across systems as a structural consequence rather than a separately engineered concern, connecting this work to existing literature distinguishing ontologies from narrower business semantic layers.

I further argue that this same structural property has a direct, quantifiable economic benefit for organizations operating large language models at scale because a resolved, five-domain Authority Relation allows a system to retrieve a small number of precise, structured facts rather than large volumes of unstructured context the model must itself re-derive meaning from, the framework supports the kind of context compression already shown, in ontology-grounded retrieval literature, to reduce token consumption and inference cost while improving factual accuracy, directly counteracting the now well-documented performance degradation large language models exhibit as context length grows. Although the motivating case throughout is the Software AI agent, I show the Authority Relation primitive transfers without modification to Physical and embodied AI, Robotics, Autonomous Vehicles, and Industrial Automation, where unresolved authority divergence carries direct physical safety consequences rather than only financial or reputational ones.

I provide a concrete implementation sketch, a JSON-LD schema, a causal-graph treatment of authority provenance, and a worked knowledge-graph query pattern to show the primitive is buildable on existing infrastructure rather than only formally specifiable, and I show this integrates directly with the Model Context Protocol and current Agentic-Memory architectures, requiring no bespoke runtime customisation.

This work is positioned relative to foundational organizational ontology (JLG Dietz's *"Understanding and Modelling Business Processes with DEMO"*; Hans Weigand, Paul Johannesson and Giancarlo Guizzardi's *"A Core Ontology of Organizational Policies"*, verified directly against the open workshop precursor to the journal paper), Upper ontologies for Real-world grounding (BFO; Guizzardi et al. 's *"Unified Foundational Ontology"*), and current industry treatments of enterprise ontology for AI agent grounding, including a closely related independent three-layer ontology for neurosymbolic agent grounding (Luong Tuan & Sanyal, 2026) that addresses a different question (how an agent should reason and speak) from the one this paper addresses (whether an agent's granted authority matches authority as actually practiced in an organisation).

ARF is deliberately T-shaped relative to this literature: it does not delve into any of these traditions' own formal apparatus, depending on and deferring to each for its specific domain, but it is the only framework I am aware of that stitches all of them together horizontally into a single, cross-domain authority relation, i.e., the connective tissue each individual tradition needs but does not itself provide. I further argue, drawing a structural analogy to Weber's account of power and "imperative control" (*Herrschaft*), that the probability, that a command will actually be obeyed, is as distinct from the formal right to issue it, that the deployment of agentic AI does not eliminate organizational power dynamics but relocates the primary site of leverage to whoever controls an agent's objective function and permission boundaries. An Authority Relation that must itself be modelled, measured, and governed using the same apparatus proposed here.

I propose a three-instrument empirical methodology for estimating the DNA-Coefficient (decision-log divergence analysis, organizational network analysis, and structured elicitation) and a fourth instrument specifically for auditing agent-design authority. I present these as a research and practice agenda rather than a validated result, and propose a structural mechanism requiring independent business, technical, and governance affirmation before any resulting score is considered valid, since each function can verify a distinct failure mode the others cannot. I conclude with a discussion of the governance, regulatory, and audit implications for enterprises currently scaling agentic AI deployment.

## 1. Introduction

Agentic AI is no longer a discrete technology initiative; it is becoming the dominant axis of enterprise IT and innovation spending. The International Data Corporation (IDC) projects Agentic AI will drive the majority of IT budget expansion over the next five years, exceeding 26 percent of worldwide IT spending and reaching 1.3 trillion dollars by 2029, and Gartner named agentic AI oversight its top cybersecurity trend for 2026, warning that unmanaged AI agent proliferation is already outpacing the governance structures meant to control it.

Set against this backdrop are poor return on investment statistics; Forbes Research's 2025 AI Survey found that fewer than one percent of executives report AI return on investment of 20 percent or greater, only three percent report 10 to 20 percent, and 53 percent report returns limited to 1 to 5 percent; separately, In June' 2025, Gartner reported that it expects more than 40 percent of Agentic AI projects to be cancelled before 2027. This enormous gap, with accelerating spend and a documented governance shortfall relative to that spend compounded by returns that remain stubbornly thin for all but a small minority of enterprises is, I argue, the central business problem this paper addresses, and it is deliberately not framed here as an engineering problem to be delegated to only the technology function.

A Persistent and Underappreciated reason that Enterprise AI initiatives and Agentic AI deployments in particular fail to scale is not solely an issue of model capability or data architecture but the absence of a shared, traceable account of who is permitted to do what, expressed consistently across every layer at which the enterprise represents itself. This issue by its very nature is tripartite:

1. Finance
    a. Is the cost/benefit ratio justified?
2. Compliance
    a. Are agentic actions consistently explainable and auditable?
3. Growth and Profitability
    a. Confidence score in capturing the productivity gains and value generated?

I reiterate this point with a concrete, structural proposal in Section 5.4, that treating semantic and authority grounding as solely a CIO or CDIO responsibility is itself part of why so many enterprises are failing to convert agentic AI spend into return, and that the framework proposed here requires by construction, not by aspiration, joint ownership across business, technical, and governance functions for exactly this reason.

Industry analysis in 2025–2026 has converged on a closely related diagnosis from a different angle; model limitations are frequently not the binding constraint on AI initiative success; the absence of a shared, machine-readable account of business meaning is [EY, 2026; McKinsey, 2026]. I extend this diagnosis one step further. Meaning alone is not sufficient. An AI agent capable of taking autonomous action needs not only a consistent definition of a "customer" or a "shipment," but a consistent, traceable, and current account of whether it or the human role it may be acting on behalf of actually holds the right to act on that customer or shipment in the present context. That account does not currently exist in a form any single enterprise ontology effort fully provides.

This gap has a specific mould.
Authority and Decision rights are typically modelled, as artifacts of a single domain: an entry in an organisation chart, a role in a RACI matrix, a permission object in an identity and access management system, or in the most rigorous academic treatments, a delegation or ruling pattern within a foundational ontology of organizations [Weigand, Johannesson, and Guizzardi, 2026]. Each of these is necessary. None, on its own or in simple combination, resolves the harder problem: that the same authority relation simultaneously exists as;

- A Social Fact (a person's actual standing and influence, which may diverge from their formal role)
- A Business Fact (what the relevant terms mean in the operational vocabulary of the domain)
- A Process Fact (whether and how the relation is codified into an auditable, compliant procedure)
- A Machine Fact (whether a system can/will actually enforce it) and
- A Real-World Fact (what the relation actually changes in the world).

When these five representations of one underlying relation are allowed to drift independently as they routinely do, because they are built and governed by different teams on different timelines the result is the familiar enterprise pathology of governance that exists on paper but is not enforceable in practice. Agentic AI does not create this pathology; it removes the human judgment that has, until now, quietly absorbed and compensated for it. A second, related benefit of cross-domain resolution deserves separate treatment, because it is conceptually distinct from the authority question and has its own substantial prior literature.

**Semantic Interoperability**, the property that different systems and agents share not merely compatible data formats but compatible meaning for the entities those formats describe has long been recognized as a harder and persistent problem than syntactic integration, requiring a shared, formal conceptualization rather than a common file format alone. Recent survey work on ontology interoperability techniques identifies design patterns, ontology matching and versioning, and data-driven validation as the dominant approaches to this problem, treating an ontology as a formal, explicit specification of a shared conceptualization in the now-standard sense established in the knowledge-representation literature [Survey: An Ecosystem for Ontology Interoperability, 2025].
A complementary, narrower industry practice distinguishes full ontologies of this kind from business-intelligence-oriented semantic layers, which solve a more limited problem of consistent metric definition within a centralized governance layer rather than full relational and inferential structure.

My Business domain (Section 3.1) corresponds most closely to what this literature calls a semantic layer. The Cross Domain resolution this framework proposes is closer to full ontological grounding, extended specifically to bind that grounding to the authority relation that governs each entity, not merely its definition. This matters concretely for a second reason beyond meaning-consistency: enterprise systems are not static. Entities change state continuously, a shipment moves from in-transit to delivered, an invoice moves from approved to paid, a customer record moves from prospect to active account and each state change is itself an event that other systems, including AI agents, must observe and react to correctly.

A long-standing strand of enterprise architecture work addresses this as a semantic event propagation problem, ensuring that when an entity's state changes in one system of record, that change is synchronized, with its meaning intact, across every other system that depends on it, rather than relying on brittle, point-to-point API integration that frequently loses or distorts meaning in transit [Cf. J McGovern, et al, 2006 "*Enterprise Service-Oriented Architectures*", and Change-Data-Capture literature].

A Five Domain framework with Object resolution as an explicit requirement (Section 3.2) directly supports this because the same Object must resolve consistently across business, process, machine, and real-world representations by construction, a state change recorded in one domain (say, the real-world fact of a shipment's delivery) propagates with its meaning preserved into the process domain (the delivery-confirmation step it satisfies) and the machine domain (the system event it should trigger), rather than requiring a separate, ad-hoc integration to be built and maintained for each pairwise connection between systems. I treat this state-change-propagation property as a second, distinct payoff of the framework, alongside, not subordinate to, the authority-resolution argument that is this paper's primary focus and one with direct relevance to the current industry push toward agent runtimes (e.g., cloud-provider agent orchestration platforms) and regulatory regimes (e.g., the EU AI Act's traceability and auditability requirements) that explicitly reward systems capable of demonstrating this kind of consistent, traceable state management at scale.

I propose a framework; **The Authority Resolution Framework, ARF**, organized around a single formal move; treating authority as a relation that must resolve across domains, rather than a property native to any one of them, and introducing an explicit term — the **DNA-Coefficient** — that captures the degree to which an organization's documented authority structure diverges from how authority is actually exercised. I use "DNA" deliberately and informally to name the idea that every organization has a unique, nuanced, and partially documented configuration of who actually decides and who actually influences those decisions, shaped by specific individuals rather than fully determined by formal structure, and that this configuration cannot be assumed to generalize across organizations of different size, sector, or culture.

This Paper makes eleven contributions:

1. A Formal Definition of the Authority Relation (AR) as a six-element primitive spanning five enterprise ontology domains (Section 3).
2. An Explicit Positioning of this primitive relative to the closest existing foundational work in organizational ontology, clarifying what this framework depends on vs what it adds (Section 2).
3. A Demonstration that the same cross-domain Object-resolution requirement which grounds the AR primitive also yields semantic interoperability and consistent state change propagation as a structural consequence, not a separate engineering effort (Section 3.4).
4. A Quantified Account of the economic benefit of this same structural property for organizations operating large language models at scale, via context compression and reduced token consumption (Section 3.5).
5. An Explicit Extension of the framework to physical and embodied AI, showing that the Authority Relation primitive transfers to robotics, autonomous vehicles, and industrial automation without modification, where unresolved authority divergence carries direct physical-safety consequences (Section 3.6).
6. A Concrete Implementation Sketch — a JSON-LD schema, a causal-graph treatment of the Justification-Chain via the W3C PROV vocabulary, and a worked graph query pattern, showing the Authority Relation primitive is buildable on existing knowledge-graph infrastructure, not just formally specifiable (Section 3.7).
7. A Demonstration that this implementation integrates directly with the Model Context Protocol and current agentic-memory architectures, requiring no bespoke runtime setup and fitting inside the integration layer enterprises are already standardizing on (Section 3.8).
8. A Proposed 3+1 instrument methodology for empirically estimating the DNA-Coefficient including a dedicated instrument for auditing who controls an AI agent's design and permission boundaries; a governance concern I argue is currently under-modelled (Section 4).

9. A Structural Mechanism requiring independent business, technical, and governance affirmation before any DNA-Coefficient score is considered valid for use, intended to make tri-functional co-ownership a procedural requirement rather than an aspiration (Section 4.5).
10. A 6-level Authority Maturity Model offered as a practical self-assessment tool for enterprises adopting this framework incrementally (Section 4.6).
11. A Discussion, grounded in a structural analogy to Weber's account of power and imperative control, of how agentic AI deployment relocates rather than removes organizational power dynamics, and what this implies for enterprise governance overall (Section 5).

I am explicit throughout about the limits of this work.
The DNA-Coefficient methodology is proposed, not validated; an empirical pilot is planned and has not yet been conducted. My positioning relative to the closest prior work is now verified for two of three related papers in full: Weigand, Johannesson, and Andersson's 2024 workshop paper and Weigand, Johannesson, and Guizzardi's 2024 Practice of Enterprise Modelling (PoEM) chapter. The final 2026 journal version by the same three authors has been confirmed to exist with the bibliographic and abstract-level detail reported here, but has not been obtained in full text, remaining paywalled at the time of writing. The Positioning against it should be read accordingly (as a provisional claim requiring further verification). I regard this as a research and practice agenda intended to be tested, critiqued, and extended by others, consistent with the cumulative, citation-based norms of the foundational ontology and enterprise architecture communities this work seeks to join.

## 2. Related work and Positioning

The literature relevant to this paper is extensive but each piece of research formalizes one piece of the enterprise well and stops at its boundary. Organizational ontology formalizes how authority is delegated and specified. Upper Ontology formalizes how entities are grounded in the real world. Industry Practice formalizes how business meaning is made consistent for AI systems to consume. AI governance frameworks formalize what controls an organization must implement to manage risk and satisfy regulation. In my view, none of these traditions asks the question this paper is built around, the lived, nuanced reality of who actually holds and wields authority in a human hierarchy. The Authority Resolution Framework (ARF) makes that nuance visible and measurable as the missing precondition for everything these other literary perspectives already do well.

### *a. Organizational ontology.*

JLG Dietz and Mulder's Enterprise Ontology and DEMO model organizations as networks of communicative acts and remain the right grounding for ARF's Social Domain. The most rigorous treatment of authority specifically comes from a sequence of three related papers by Weigand and Johannesson, with Andersson and then Guizzardi as successive co-authors.
Weigand, Johannesson, and Andersson's VMBO 2024 workshop paper develops two Ontological patterns: an organizational policy pattern and a policy document pattern. Weigand, Johannesson, and Guizzardi's PoEM 2024 chapter, "The Dual Nature of Organizational Policies," explicitly builds on and extends the workshop paper, and developing two further patterns: A Delegation/Authorization pattern and a Ruling pattern for four patterns in total.

The Organizational policy pattern's worked example is structurally identical to the one I develop (Section 3.3), a Treasurer role with the duty and power to pay invoices, requiring departmental head consent above a $500 threshold. The Policy document pattern's own ERP/Treasurer illustration briefly notes that the IT system interpreting a rule should, strictly speaking, be distinguished from the rule itself — an observation adjacent to ARF's Machine domain that is not extended into a mechanism for detecting when system and rule diverge in practice. The Delegation/Authorization pattern models a 3-party structure (Customer, Provider/Principal, Agent) in which a delegated action may only partially discharge the principal's underlying obligation to a third party, illustrated with an example in which delegating one task does not fully satisfy a broader service obligation if a further step (such as delivery) is also required.

ARF's current Authority Relation primitive, built around a single Object per relation, does not yet capture this kind of partial, multi-step obligation, and I flag this as a genuine strength of their treatment and an open extension for ARF. The ruling pattern grounds authority's ultimate origin in a reciprocal exchange between a governing party and a community, and argues that a policy's effectiveness depends on whether it is actually acted upon, not merely believed to exist, a position close in spirit to ARF's premise that documented authority only matters insofar as it is exercised, but framed as a general claim about what makes any policy effective in principle, rather than as an instrument for measuring divergence in a specific organization at a specific time.

ARF's DNA-Coefficient is the latter; a scored, directional, per-instance measurement, not a general claim about policy effectiveness. ARF depends on these patterns directly for how an Authority Relation's documented half should be formalized in the social domain. It does not compete with them. The authors' own stated future work is directly relevant here, their PoEM chapter's conclusion states that the policy ontology requires further empirical validation against an actual, real-world policy, and the workshop precursor separately flags cross-level "policy grounding" as work still to be done. Neither gap is the one ARF addresses, but both confirm the same underlying limitation from the authors' own account; a policy that is internally well-formed and ontologically well-specified by their patterns is not thereby empirically verified to be the policy currently, accurately practiced.

A further journal version, "A Core Ontology of Organizational Policies" [Weigand, Johannesson, & Guizzardi, 2026, Software and Systems Modelling], appears to extend this same four pattern structure with additional validation against a university domain and the SBVR standard. I have confirmed its abstract and structure but have not obtained its full text, which remains paywalled, and flag this as the one remaining unverified item in this lineage. The earliest, a VMBO 2024 workshop paper by Weigand, Johannesson, and Andersson, introduces two initial patterns (organizational policy, policy document) and the core tension the later work resolves; UFO's existing treatment of a policy as a "normative description" is incompatible with UFO-L's treatment of legal positions as agent-relators.

The direct successor, "The Dual Nature of Organizational Policies" [Weigand, Johannesson, & Guizzardi, PoEM 2024, Springer LNBIP vol. 538], resolves this tension by arguing policies have a dual nature; simultaneously an institutional relator (rights and duties between a Governor and a Policy Subject, in UFO-L terms) and an artefactual policy document (with its own make-plan, use-plan, and lifecycle, including an Activation event) and develops this into four full ontological patterns; Organizational policy, Policy document, Delegation/Authorization, and a Ruling pattern addressing how a Principal's power over a community is itself grounded in power delegated by that community, including the power to sanction and to create new policies.

The Organizational policy pattern's worked example is structurally identical to the one I develop in Section 3.3; a Treasurer role with the duty and power to pay invoices, requiring written department-head consent above a $500 threshold. ARF adopts this pattern directly for how an Authority Relation's documented half should be formalized; it does not compete with it, and the dual-nature distinction between a policy as institutional relator and its policy document as artifact maps closely enough onto my own distinction between the Justification-Chain (the institutional basis) and the Process/Machine domains (the artifact's operational instantiation) that an implementation of ARF should treat the two as compatible rather than parallel inventions.

The authors' own ERP/Treasurer illustration of the policy document pattern notes, in passing, that "the IT application interpreting the rules" should be distinguished "from the rules themselves," an observation directly adjacent to ARF's Machine domain but, as in the workshop precursor, this is not developed into a mechanism for detecting when the two diverge in practice; the worked university-domain application in their paper's Section 4 validates that the ontology can represent a range of real organizational policy statements correctly, not that any of those statements still match current practice. The authors state their own future work plainly: the policy ontology "requires more empirical validation, for instance by applying it to an actual policy," and they aim next to connect it to a separate decision-making ontology, neither of which is the documented-versus-practiced divergence question ARF addresses, though both point to the same underlying limitation their paper does not close.

an ontology that can correctly represent what a policy specifies is not thereby validated against whether that specification still holds in practice. That is the specific layer ARF adds.

A further extended journal treatment, “A core ontology of organizational policies” [Weigand, Johannesson, & Guizzardi, 2026, Software and Systems Modelling], appears from its abstract to develop the same four-pattern structure already verified above, evaluated additionally against the SBVR business rule standard; I have confirmed this paper’s abstract and structure but not its full text, which remains paywalled, and flag this as the one remaining unverified item in this lineage — though, having now read the PoEM chapter in full, I consider it unlikely that the journal version’s additional SBVR evaluation changes the positioning established here.

### b. Upper Ontology.

BFO and Guizzardi et al. ‘s UFO (2022) provide the rigorous, domain independent grounding ARF’s Real-world domain depends on for representing physical and operational fact. ARF does not extend this apparatus; it uses whichever upper ontology an enterprise already relies on and requires only that the Authority Relation’s Object resolve consistently against it.

### c. Industry practice and Explainability Research.

A fast-growing body of consulting and vendor work treats ontology as a “control plane” binding business meaning to AI consumable structure [EY, 2026; CIO.com, 2026], correctly diagnosing that ungrounded meaning produces ungrounded agent behaviour. Confalonieri and Guizzardi (2025) make the related, more rigorous academic case that ontological grounding supports explainable, hallucination-resistant AI reasoning. Both strands establish that ontological grounding is now a credible, mainstream mechanism for constraining AI behaviour, a useful precedent for ARF’s agent-design-authority instrument (Section 4.3), but neither asks the authority question, this paper is built around.

**The Closest parallel**. Luong Tuan and Sanyal (2026) propose a three-layer ontology (Role, Domain, Interaction) for grounding enterprise LLM agents, validated through a real 1,800-run experiment — the most empirically rigorous work in this space to date, and a capability ARF does not yet have. Their framework optimizes how an agent reasons and speaks; ARF verifies whether the authority it’s been given is still real. The two are complementary by construction - An enterprise could run their architecture for context assembly and ARF for authority verification without either subsuming the other. I discuss their economically relevant findings, including a result that usefully qualifies my own claims, in Section 3.5.

### d. AI Governance and Compliance.

Eisenberg, Gamboa, and Sherman’s (2025) Unified Control Framework, from Credo AI, is the most operationally mature governance artifact I am aware of — fifty risk scenarios, forty-two implementable controls, validated against the Colorado AI Act, with concrete tooling guidance ARF does not currently offer. UCF answers what an organization must do to manage a risk or satisfy a regulation. It does not, and by its authors’ own stated limitation cannot, verify whether a correctly specified control is being exercised by the person or function it names — independent verification of faithful implementation is explicitly outside UCF’s scope. A UCF control can be perfectly mapped to the right regulation and silently unenforced by anyone the documentation recognizes.
That is precisely the failure mode ARF’s DNA-Coefficient is built to catch. The two frameworks are not competitors; an enterprise serious about agentic AI risk needs both, Comprehensive control coverage from something like UCF, and Verified authority underneath every control, human or agentic, via ARF.

### e. Theoretical grounding.

ARF claims that organizational authority is only partially formalizable and draws on four traditions to make that claim precise rather than convenient.

1. Aristotle’s Metaphysics, Book Γ (Gamma), grounds the requirement that an Authority Relation be consistent across domains in the same principle of non-contradiction that any formal reasoning system, human or machine, already depends on.

2. Daniel Kahneman's System 1/System 2 distinction explains why human authority routinely runs on fast, intuitive judgment that a purely formal machine representation cannot natively capture. This is the reason the DNA-Coefficient is a calibrated adjustment, not a claim to fully formalize trust or culture.
3. Noam Chomsky's Universal Grammar supplies a structural image rather than a theoretical commitment, a shared deep structure recoverable beneath surface variation, as a way of motivating why ARF looks for one Authority Relation beneath five differently-worded domains, and the analogy does not depend on Universal Grammar being correct as linguistic theory, a contested claim I take no position on, since these large language models acquire fluent language behaviour through statistical exposure rather than an innate grammar module. Readers who find the linguistic theory itself unpersuasive can set the analogy aside without any loss to the formal argument in Section 3, which depends on the Authority Relation primitive directly, not on this.
4. The Organizational-Anthropology tradition (Douglas, Geertz, Graeber) grounds why that structure must be measured per organization rather than assumed from a template.

The throughline across all five comparisons above is the same. Every serious prior effort formalizes a real piece of the enterprise and stops at its boundary. But the deeper issue is not that authority has been split across separate literatures — it is that the hierarchy itself was never the clean, formal structure an org chart implies. Command-and-control in a real organization is a deeply nuanced, lived thing: tenure, trust, unwritten deference, the senior person everyone still quietly calls long after the title has moved on. Humans run on that nuance instinctively and have for as long as organizations have existed. Agentic AI is the first actor in the history of these institutions that cannot perceive it at all — not poorly, not partially, but structurally not at all, because nuance of that kind has never been written down anywhere a system could read it. ARF's contribution is not a sixth literature filling a sixth gap. It is the claim that this invisible nuance has to be made visible and measurable before any of the other five literatures' careful work can be safely handed to something that cannot sense it the way a person does.

## 3. Formal Framework

### 3.1 The Five Enterprise Ontology Domains

I define five domains in which an enterprise represents itself, summarized in Table 1.
These domains are not proposed as novel in themselves, each has an established grounding tradition but their explicit enumeration as a closed, named set is necessary scaffolding for the cross-domain resolution mechanism proposed in Section 3.2.

Table 1: The Five Enterprise Ontology Domains

| Domain | Definition | Grounding Tradition | Boundary |
|---|---|---|---|
| Social | Formal roles, titles, mandates, and hierarchy, together with the informal, only partially documented reality of who actually exercises influence over a given decision. | DEMO; UFO/DOLCE Organizational Ontology. Organizational Anthropology | A documented role holder lacks the actual standing to exercise the authority nominally assigned to them, or vice versa. |
| Business | The operational vocabulary of the enterprise; concepts, events, entities, terms, acronyms, and abbreviations that enable efficient communication. | Business Glossaries; SBVR. Domain-driven design | The same term denotes different concepts in different functions (e.g., "revenue" recognized under different bases in Finance vs Sales) |
| Process | Business processes once codified into standardized, auditable, legally compliant procedures. | BPMN and Process Mining Compliance and Audit frameworks | Tacit practice diverges from Documented procedure, and the divergence is invisible until an audit or incident surfaces it. |
| Machine | The formal, executable representation of entities, relationships, and Permissions. | Code, Scripts, Logs, Version controlled artifacts, Knowledge graphs RDF/OWL; Software Engineering, including Object-oriented design patterns (inheritance, polymorphism, class hierarchies). | Permissions are encoded and enforced but not traceable to an accountable human decision or current organizational reality. |
| Real-World | The operational and physical state of the world the enterprise must track and respond to but do not fully control, including externally sourced knowledge via foundation models. | Upper ontologies (e.g., BFO) for physical/operational grounding; Foundation-model mediated external knowledge as a distinct sub-case. | Systems and agents reason over an Internal representation that has silently diverged from current ground Truth. |

I do not claim these five domains are exhaustive of every way an enterprise might be ontologically described, nor that the boundaries between them are perfectly sharp, Process and Business Vocabulary, for instance, plainly interact. I claim only that this five-way decomposition is fine-grained enough to expose where authority relations are typically allowed to drift independently across domains in current practice, which is the specific failure mode this framework addresses.

## 3.2 The Authority Relation as a cross-domain primitive

I define an Authority Relation (AR) — the central formal object of ARF — as a six-element tuple.
*AR = (Actor, Action, Object, Domain-Context, Justification-Chain, DNA-Coefficient)*

***Actor***. A Role or agent capable of holding authority, defined within the social domain.
This includes human role-holders and, critically for the agentic AI case this paper motivates, non-human agents holding authority by delegation from a human or institutional principal. I follow the organizational-ontology tradition in treating delegated authority as a first-class relation rather than collapsing the delegate's authority into that of the delegator; an agent's authority is its own AR instance, traceable to but distinct from the authority of whoever configured it.

***Action***. A Permitted or required act, drawn from a controlled vocabulary defined in the business domain (e.g., approve, escalate, override, delegate, veto). I treat the Action element as the natural locus for the object-oriented analogy introduced in Section 2.4: the same Action (e.g., "approve") behaves polymorphically depending on Domain Context, and specific role-actions can be modelled as inheriting baseline permissions from a more general class (e.g., a "Regional Director" role inheriting from a generic "Approver" class, with explicit overrides).

***Object***. The Entity being acted upon. The Object is the element that must resolve consistently across the business, process, machine, and real-world domains — i.e., the "customer," "shipment," or "price" referenced by an Authority Relation must denote the same entity in the business glossary, the codified process, the machine's permission system, and the real-world state, or the relation cannot be said to be coherently specified at all. This requirement — Object resolution across domains — is the primary binding mechanism by which this framework forces what would otherwise be four or five independently governed representations into a single, checkable relation.

***Domain-Context.*** The Bounded scope within which the authority holds — a jurisdiction, business unit, system boundary, or similar. Authority is treated as inherently scoped rather than universal; an AR instance without an explicit Domain-Context is underspecified.

***Justification-Chain.*** The Traceable basis for the authority — a governance policy, a specific delegation event, a regulatory requirement, or a documented business decision. This element is what makes an AR instance auditable, and what allows a human or an AI agent to produce an explanation for why a given action was or was not permitted, rather than treating authorization as an opaque permission check.

***DNA-Coefficient.*** A Calibrated, empirically estimated measure of the divergence between the AR instance as formally specified (via the preceding five elements) and how it is actually exercised in practice within a specific organization. This is the genuinely novel element of the framework relative to existing organizational-ontology and policy-ontology treatments, and is developed fully in Section 4.

## 3.3 Worked illustration

I deliberately choose an illustration structurally close to the worked example in Weigand, Johannesson, and Andersson (2024) — a role-based approval threshold with escalation above a monetary amount — because it is close to a canonical example in this literature, and because using a recognizably similar structure makes the contrast with their treatment precise rather than asserted. Their example: a Treasurer role authorized to pay invoices, requiring departmental head consent above a $500 threshold, with their organizational policy pattern formally capturing the role, the delegation from the Corporation, and the resource (invoice payment) the policy governs.

I extend this with a structurally identical case from a different domain, then show concretely what their pattern does not, and is not intended to, capture: whether the documented policy still matches who actually exercises that authority in practice. To ground the formalism, consider a recurring enterprise decision: approval of a price discount exceeding ten percent of list price, in a consumer-packaged goods business operating across multiple national markets. I develop this example in more detail than a single AR instantiation in order to show concretely how the framework's elements interact, and how the DNA-Coefficient instruments from Section 4 would apply to it in practice.

**The Formal Specification.** A multinational CPG enterprise's pricing governance policy states: "discounts up to ten percent of list price may be approved by a Country Commercial Manager; discounts above ten percent require approval from the Regional Commercial Director; discounts above twenty-five percent require Regional Commercial Director approval plus finance sign-off".

This is the documented Authority Relation:
***Actor*** *= Regional Commercial Director (role).*
***Action*** *= Approve.*
***Object*** *= A Discount request exceeding the ten-percent threshold.*
***Domain-Context*** *= The Specific country or cluster of markets the Director covers.*
***Justification-Chain*** *= The Enterprise's pricing governance policy document, version-controlled, dated.*

**What is already well-covered by existing formal work.**
Everything in the preceding paragraph — the role, its delegation from the Corporation (here, the enterprise), the threshold-based escalation, and the policy document's status as a regulatory artifact distinct from the policy it instantiates is precisely what Weigand, Johannesson, and Andersson's organizational policy pattern is built to formalize, in exactly the structure their Treasurer example demonstrates. An implementation of this framework's social domain should use their pattern directly for this part, not reinvent it. Their formalism, applied here, would correctly model the Regional Commercial Director role as holding a delegated legal position over the discount-approval action, scoped to the relevant organizational resource (pricing), with the policy document as the artifact specifying the threshold. A system built only on their pattern, however well implemented, would stop here. It is not intended to verify, or make any further claim, about whether this formally correct specification matches what actually happens in the organization.

**Where the five domains can silently diverge — the gap their pattern does not address.**
In the Social domain, the Regional Commercial Director role may be held by someone newly promoted into the position, while the previous incumbent — now in an adjacent role with no formal pricing authority at all continues to be the person sales teams actually escalate to, because of accumulated trust and historical relationships with key accounts.

In the Business domain, "discount" itself may be defined inconsistently; Sales may calculate the percentage against list price, while Finance calculates it against net invoice price after existing standard terms, producing different answers as to whether a given deal even crosses the ten-percent threshold.

In the Process domain, the codified approval workflow may require the discount request to be logged in a specific system before approval but in practice, urgent deals are frequently approved verbally or over email first, with the system entry completed retroactively, meaning the auditable process record does not reflect the actual sequence of authority being exercised.

In the Machine domain, the CRM or pricing system's permission configuration may still reference the previous Regional Director by name or by a role mapping that was never updated after the promotion, meaning the system would technically block the new incumbent from approving a deal the policy says they are authorized to approve.

In the Real-world domain, the discount's actual commercial consequence, its effect on realized margin for that account over the contract period may not be visible to the approver at the point of decision at all, since margin reporting typically lags the transaction.

**Applying the DNA-Coefficient instruments.**

1. A **Decision-Log Divergence Analysis**, sampling the past twelve months of discount approvals above the ten-percent threshold in this market, would compare the system's recorded approver against the actual approval trail reconstructed from email and CRM notes. If the previous incumbent is found to have driven the substantive approval decision in a material share of cases despite no longer holding the formal role, this registers as a measurable, directional divergence, i.e., Authority sitting with an individual rather than the documented role.
2. An **Organizational Network Analysis** of the sales and commercial function in that market, using meeting and correspondence metadata, would likely show the previous incumbent retaining unusually high centrality relative to their current formal position, corroborating the decision-log finding through an independent method.
3. **Structured Elicitation** — Short interviews with country sales managers asking who they would actually consult on a borderline discount decision would surface the same name even among staff who have never seen a decision log, capturing the System-1-type trust effect directly.

Where all three instruments converge on the same finding, the composite DNA-Coefficient for this AR instance would be scored as a high-magnitude, clearly directional divergence, flagging it as a priority for governance attention specifically because it is both highstakes (material to margin) and high-divergence, precisely the combination a generic, formally-correct-on-paper permissioning scheme for an agentic AI pricing assistant would miss entirely, since such a system would be configured against the documented role mapping in the machine domain, not against the lived reality the three instruments above reveal.

**The Applied Stakes.**
If this enterprise were to deploy an agentic AI system to triage or pre-approve discount requests, a realistic and increasingly common application in CPG commercial operations; Configuring that agent's permissions against the machine domain's role mapping alone would produce a system that is technically compliant with documented policy and simultaneously disconnected from how the organization actually makes this decision. The agent might correctly block the new Regional Director from a system perspective while having no mechanism to recognize that the actual decisionmaker the organization relies on is someone else entirely, with no formal authority in the system at all.

This is the concrete, operational version of the abstract claim in Section 1 that an authority relation not resolved across all five domains cannot be safely delegated to a non-human actor, because the agent inherits whichever domain's representation it was configured against, and has no native way to detect that the other four domains disagree. This illustration is constructed for expository purposes and is not drawn from a single real organization's data, though each element reflects a documented and common pattern in enterprise pricing governance and role-transition literature. Section 4.4 discusses the empirical work required to move from illustration to validated instrument.

### 3.4 Semantic Interoperability and State-change Propagation as Structural Consequences

The Object-resolution requirement introduced in Section 3.2 that the entity referenced by an Authority Relation must denote the same thing across the Social, Business, Process, Machine, and Real-World domains was motivated above primarily by the authority problem. It is worth making explicit that this same requirement yields two further, independently valuable properties as a structural consequence, not as separate work that must be additionally engineered.

**Semantic Interoperability.**
A substantial existing literature distinguishes ontologies proper — which provide explicit relationship types, support for logical inference, and standard interoperable representations such as RDF, OWL, and SKOS — from narrower business-intelligence-style semantic layers, which solve consistent metric definition within a centralized governance layer but do not support the same depth of relational inference. My Business domain (Table 1) corresponds most closely to a semantic layer in this sense.

Because my framework requires the Object referenced in an Authority Relation to resolve consistently across the Business domain and the Social, Process, Machine, and Real-world domains simultaneously, an enterprise adopting this framework acquires, as a direct consequence of doing so, the cross-system semantic consistency that a semantic layer alone does not provide and that a full ontological treatment is specifically designed to deliver, without requiring a separate semantic-interoperability initiative to be undertaken on top of the authority-governance work motivating this paper.

**State-Change Propagation.**
Enterprise entities are not static. A Shipment transitions from in-transit to delivered, an invoice from approved to paid, a customer record from prospect to active. Each such transition is an event that other systems and, increasingly, autonomous AI agents must observe and act on correctly. This is conventionally treated as an integration problem, addressed through point-to-point APIs, event buses, or change-data capture pipelines, each requiring bespoke logic to preserve meaning as state changes propagate between systems built on different internal representations.

Closely related work proposes using an explicit ontology model within an event-driven architecture specifically to ensure that event-driven services across an organization and its sub-organizations follow predefined organizational eventaccess rules, with the ontology defining the domain-specific concepts attached to each event [Cao, Yang, & Deng, 2020].

This is, to my knowledge, the closest existing academic precedent for binding event and state-change semantics to organizational access rules via an explicit ontology; it does not, however, address the authority-divergence problem — documented versus lived practice — that is this paper's primary focus, nor extend the binding across business, process, and real-world domains simultaneously as I propose.

A Framework in which the same Object is required to resolve consistently across all five of my domains by construction extends this logic and reduces propagation to a structural property rather than a per-integration engineering task: a state change recorded as a real-world fact (the shipment's delivery) propagates with its meaning intact into the corresponding process-domain event (the delivery confirmation step it satisfies) and the corresponding machine-domain event (the system trigger it should fire), because the Object underlying all three was never permitted to diverge in the first place.

I note that this property is most useful precisely where it intersects with the authority question motivating the rest of this paper: an AI agent that correctly observes a state change but cannot correctly determine whether it currently holds the authority to act on that change has not been helped by interoperability alone, which is why I treat the two properties; Consistent meaning and Resolved authority as jointly necessary rather than presenting either as sufficient on its own.

I note, finally, that this property is increasingly relevant to current regulatory developments; Traceability and Auditability requirements in AI-specific regulation (e.g., provisions within the EU AI Act applicable to high-risk AI systems) are structurally well-served by an architecture that can demonstrate, by construction, that a given state change and the authority to act on it were both consistently and traceably resolved, a claim considerably harder to support credibly under a point-to-point integration architecture with no shared cross-domain primitive.

### 3.5 Economic consequences- Context Compression and Token Cost

A Separate, more directly quantifiable benefit follows from the same structural property and deserves explicit treatment, since it speaks to enterprise adoption incentives independent of the governance argument that is this paper's central concern. Large language models exhibit well-documented performance degradation as input context length increases: Du et al. (2025) show that even with perfect retrieval, LLM performance degrades 13.9 to 85 percent as context length increases, demonstrating that context volume itself imposes a cost independent of content quality.

This finding is corroborated by Luong Tuan and Sanyal (2026), introduced in Section 2.3 as the closest existing parallel to the present work, a separately constructed three-layer ontology addressing agent reasoning and speech rather than authority, but one whose empirical findings on context efficiency and ontology-versus-retrieval performance are directly relevant here regardless of the difference in the two frameworks' core purpose. Their controlled 1,800-run evaluation across five regulated industries and three LLM architectures finds statistically significant improvements in metric accuracy and role consistency from ontological grounding, with effects strongest precisely where LLM parametric knowledge is weakest. An effect they term the *inverse parametric knowledge effect*.

I note their finding that retrieval-augmented generation using curated, well-organized unstructured text can be competitive with structured ontological injection on pure terminology recall, while structured ontology provides a measurable advantage specifically on relational and role-consistency tasks. This is a useful, empirically grounded caution against overclaiming uniform ontology superiority, and I adopt the more modest, qualified framing it supports throughout this section.

Retrieval-augmented generation systems grounded in an explicit ontology address the context-volume cost directly by retrieving a small number of precise, structured facts, entities and their relationships, materialized as compact triples or hyperedges rather than retrieving larger volumes of unstructured passage text that the model must itself parse and re-derive meaning from.

Sharma, Kumar, and Li (2025) demonstrate this concretely; their OG-RAG system, which constructs a hypergraph representation of domain documents grounded in domain-specific ontology and retrieves a minimal covering set of hyperedges per query, increases the recall of accurate facts by 55 percent and improves response correctness by 40 percent across four different LLMs relative to standard retrieval, while enabling 30 percent faster attribution of responses to their supporting context.

I do not present new empirical results of my own on this point. I note it because the mechanism generalizes directly to my framework. A Query resolved against a well-formed Authority Relation, for instance, "does this actor currently hold authority to approve this action on this object, in this context?" can be answered by retrieving the specific, structured tuple in question, together with its Justification-Chain, rather than retrieving and asking the model to interpret an unstructured policy document, an email thread, and a system log separately and reconcile them itself. Because the Object is required to resolve consistently across all five domains by construction (Section 3.2), the structured retrieval is also more reliable. There is, by design, less ambiguity for the model to resolve about which domain's representation of the entity is authoritative, though I note, per Luong Tuan and Sanyal's findings above, that this advantage is most pronounced for relational and authority-consistency queries specifically, not necessarily for simple terminology lookup, where well-curated unstructured retrieval can be competitive.

The Practical consequence for enterprises operating language models and agentic systems at scale is twofold, a direct reduction in token consumption and inference cost per query for authority and relationship dependent queries specifically, and an indirect reduction in hallucination risk attributable to context-length degradation, both following from the same cross-domain resolution this framework requires for governance reasons. I regard this as a genuine economic incentive for adoption, additional to and independent of the governance and risk-reduction case developed in Section 5.3, and one likely to be legible to a technology budget holder considering this framework's adoption cost even where the governance argument alone is not.

### 3.6 Extension to Physical and Embodied AI

The argument so far has used software agents and large language models as the motivating case throughout, and most of the empirical and economic grounding cited in Sections 3.4 and 3.5 is drawn from that literature specifically. This is a framing choice, not a structural limit on what ARF addresses. Nothing in the five-domain decomposition or the Authority Relation primitive depends on the Actor or the system enforcing a given Action being software rather than hardware, and 2026 has seen physical AI, AI Systems that perceive, decide, and act in the physical world through robotics, autonomous vehicles, and industrial automation move from research demonstration to commercial deployment at a pace that makes this extension necessary to state explicitly rather than leave implicit.

The shift is already underway at meaningful scale: a 2026 Deloitte enterprise survey of more than 3,200 business and IT leaders found that 58 percent of organizations were already using physical AI to some extent, with adoption projected to reach 80 percent within two years, and Singapore's Infocomm Media

Development Authority has already extended its model AI governance framework explicitly to agentic systems that control devices, recommending least-privilege permissions and human approval gates for exactly this class of system. A substantial and well-capitalized segment of frontier AI research is now organized specifically around this problem: AMI Labs, founded by Yann LeCun with a $1.03 billion seed round in 2026, is building "world models". AI systems trained on sensor and spatial data rather than text, intended to ground reasoning and planning in physical reality with stated target applications including industrial robots, autonomous vehicles, and surgical robots, domains the company itself describes as ones where reliability and safety are non-negotiable.

The Authority Relation primitive transfers to this setting with the Object reinterpreted, not redefined; where the worked illustration in Section 3.3 resolves a discount-approval decision, the equivalent Physical-AI instance resolves a movement, manipulation, or intervention decision; a robot approaching a human worker on a factory floor, an autonomous vehicle executing an unplanned manoeuvre, a surgical robot adjusting an incision path mid-procedure.

Each is, formally, the same six-element tuple:
An Actor (a role or, a delegated non-human agent) takes an Action on an Object, within a Domain-Context (a specific facility, vehicle class, procedure type), grounded in a Justification-Chain (Safety Certification, an Operating protocol, Regulatory clearance), with a DNA-Coefficient capturing whether the documented operating envelope still matches how the system is actually run in practice, for instance, whether a robot's certified safety parameters have been informally adjusted on the shop floor to hit a production target, a divergence with direct physical-safety consequences (not just financial or reputational ones).

The Stakes of unresolved divergence are correspondingly higher in this domain, a point the embodied-AI safety literature already converges on independently of ARF. Failures in physical environments affect infrastructure, property, and human safety directly, not only data integrity or financial outcomes, and the recurring question raised across recent industry and standards venues — who remains accountable when decisions are automated — is the authority question this paper addresses, asked in a setting where the cost of an unresolved answer is measured in physical harm rather than dollars.

I do not present new empirical work specific to physical AI in this paper; I note this extension because the framework's claim to generality should be stated plainly rather than left to be inferred, and because the same co-ownership mechanism proposed in Section 4.5 requiring business, technical, and governance affirmation before an authority finding is considered valid applies at least as urgently where the technical function in question is robotics or controls engineering rather than software, and the governance affirmation in question concerns physical safety rather than financial exposure.

### 3.7 Implementation Sketch- A Concrete Schema and Query Pattern

The argument so far has been deliberately implementation-agnostic, on the view that the Authority Relation's formal structure should not depend on any one storage technology. This section sketches one concrete realization, using JSON-LD and a knowledge-graph backend, to show the primitive is buildable, and to give an implementing team a starting artifact rather than only a formal definition.

**Why JSON-LD and a graph, not a relational table.**
A Relational schema would force a fixed set of columns onto an Authority Relation, but the Object resolution requirement in Section 3.2 specifically needs the same entity to be addressable from the Business, Process, Machine, and Real-world domains simultaneously, with new domain-specific attributes attachable without a schema migration each time a new system is onboarded.

JSON-LD provides this directly. Every entity carries an @id (a stable, dereferenceable identifier playing the role of the Object across all domains) and an @type drawn from a shared vocabulary, while remaining valid, parse-able JSON that any application stack can consume without graph-specific tooling.
A Knowledge-graph backend (any RDF triple store such as GraphDB, AllegroGraph, or a labelled-property graph such as Neo4j, AWS Neptune with an RDF/JSON-LD export layer) is the natural store for the resulting structure, since the Authority Relation is, at its core, a set of typed relations between nodes, not a set of independent records.

**The AR instance as JSON-LD.**

The discount-approval illustration from Section 3.3 instantiates as follows:

```
{
 "@context": "https://arf-schema.org/v1/context.jsonld",
 "@id": "arf:ar-instance:disc-2026-0614-7741",
 "@type": "arf:AuthorityRelation",
 "arf:actor": {
 "@id": "org:role:regional-commercial-director-emea-north",
 "@type": "arf:OrganizationalRole",
 "arf:heldBy": "org:person:p-15683",
 "arf:since": "2024-01-01"
 },
  "arf:action": {
 "@id": "arf:action:approve",
 "@type": "arf:GovernedAction",
 "skos:prefLabel": "Approve discount"
 },
 "arf:object": {
 "@id": "biz:discount-request:dr-88292",
 "@type": "biz:DiscountRequest",
 "arf:resolvesTo": [
 "process:workflow-step:pricing-approval-gate-3",
 "sys:crm-permission:perm-4471",
 "world:fact:margin-impact-q3-2026"
 ]
 },
 "arf:domainContext": {
 "@id": "org:market:emea-north",
 "arf:threshold": "0.10",
"arf:currency": "EUR"
 },
 "arf:justificationChain": {
 "@id": "policy:pricing-governance-v7",
 "@type": "arf:PolicyDocument",
 "arf:effectiveDate": "2025-09-01",
 "prov:wasDerivedFrom": "policy:pricing-governance-v6"
 },
 "arf:dnaCoefficient": {
 "@type": "arf:DNAScore",
 "arf:divergenceMagnitude": 0.52,
 "arf:direction": "actual-exceeds-documented",
 "arf:measuredVia": ["arf:decisionLogDivergence", "arf:structuredElicitation"],
 "arf:measuredAt": "2026-04-11",
 "arf:affirmedBy": {
 "arf:business": "org:person:p-55112",
 "arf:technical": "org:person:p-66833",
 "arf:governance": "org:person:p-81920"
 }
 }
}
```

Two design choices here are load bearing rather than incidental. First, *arf:resolvesTo* is an explicit array of cross-domain identifiers. The Process workflow step, the Machine permission object, and the Real-world fact the Object corresponds to, which operationalizes Section 3.2's resolution requirement as a checkable property. A Validator can confirm at write time that all four referenced identifiers exist and are not stale, rather than relying on convention. Second, *arf:affirmedBy* operationalizes the trifunctional co-ownership mechanism from Section 4.5 directly in the data model. The Schema makes a DNA-Coefficient record structurally incomplete, not merely procedurally discouraged, until all three affirming identities are populated, a validating system can reject a write that lacks any one of them, which is a stronger guarantee than a policy document asking teams to remember to collaborate.

**The Justification-Chain as a causal graph.**

The *prov:wasDerivedFrom* property above is not decorative. It places the Justification-Chain inside the W3C PROV provenance vocabulary, which models derivation as a directed graph of causal dependency between artifacts, agents, and activities. This matters specifically because a Justification-Chain is not just a pointer to a policy document but a claim about why the current authority specification is valid. It was derived from a prior version, by a specific process, under a specific approval.

Modelling this as a causal graph rather than a single foreign-key reference allows a query to walk backward through the chain (which prior policy version authorized this one, and who approved that derivation), precisely the audit trace a regulator or an internal audit function would need to request under the traceability requirements discussed in Section 3.4, and the same structure that would let an AI agent explain not just that it has permission, but the derivation chain establishing why.

**A Worked Query.**
An Agent or a permissioning service checking authority at runtime would issue a graph query of the following shape (expressed in SPARQL against a triplestore realization of the schema above, or expressed against a labelled-property graph using Cypher):

```
PREFIX arf: <https://arf-schema.org/v1/terms#>
SELECT ?actor ?dnaDivergence ?direction ?affirmed
WHERE {
 ?ar arf:object/arf:resolvesTo <sys:crm-permission:perm-4471> ;
 arf:actor ?actorRole ;
 arf:dnaCoefficient ?dna .
 ?actorRole arf:heldBy ?actor .
 ?dna arf:divergenceMagnitude ?dnaDivergence ;
 arf:direction ?direction .
 FILTER NOT EXISTS {
 ?dna arf:affirmedBy ?aff .
 FILTER (?aff = "" || !BOUND(?aff))
 }
}
```

This query starts from a Machine-domain permission object the agent is about to rely on, traverses backward through the Object's cross-domain resolution to the governing Authority Relation, and returns the current actor, the measured divergence, and its direction, and implicitly filters out any AR instance whose DNA-Coefficient lacks complete tri-functional affirmation. An agent receiving a non-empty, fully-affirmed result with low divergence magnitude has a defensible basis to act whereas an agent receiving a high-divergence or incompletely affirmed result has a structured, machine-readable reason to escalate to a human rather than act on a permission the data model itself flags as unverified.

**Where this connects to the economics of Section 3.5.**
This schema is also what makes the token-compression argument concrete rather than aspirational. The query above returns a handful of structured triples, actor, divergence score, direction, affirmation status rather than requiring an LLM to ingest the full text of the pricing policy document, the CRM's permission configuration export, and a thread of approval emails and infer the answer itself. This is the same mechanism Sharma, Kumar, and Li's OG-RAG system formalizes at the retrieval-architecture level (Section 3.5), applied here to the specific query shape an authority check requires.

The General economic pattern, that compressing what reaches a model's context window, rather than asking the model to parse raw volume, both reduces cost and improves reliability is not unique to ontology-grounded retrieval. It appears even at the level of raw tool-call output.
RTK ("Rust Token Killer"), an Open-source developer tool unrelated to ARF, compresses verbose shell command output before it reaches a coding agent's context window and reports 60 to 90 percent token reductions on common developer commands with no loss of information relevant to the task.
The Mechanism is syntactic (filtering, deduplication, truncation of text) rather than semantic (resolution to a single authoritative structured fact), and so it is not a substitute for the ontology-grounded compression argued for above, but its existence and reported results corroborate, from a different layer of the AI tooling stack, the same underlying economic pressure motivating this section; that the cost of an AI system's context is a real, addressable engineering variable, and that compressing what reaches the model, by whatever mechanism fits the layer in question, pays for itself in measurable token and reliability terms.

## 3.8 Integration substrate- Agentic memory and the Model Context Protocol

The Implementation sketch above is deliberately storage-agnostic, but it is worth being explicit about how an Authority Relation store would actually reach a running agent in current production architectures, since this is now a settled, rapidly maturing part of the agentic AI stack rather than a speculative integration point. The Model Context Protocol (MCP), an open standard introduced by Anthropic in late 2024 and since adopted across essentially the entire industry, including OpenAI, Google DeepMind, Microsoft, and Amazon, with tens of thousands of production server implementations by early 2026 has become the default mechanism by which an AI agent discovers and calls external tools and data sources, including persistent, cross-session memory stores. MCP's architecture is structured so that an agent's "discover" step exposes only the specific actions a given client is permitted to invoke, not the full surface of an underlying system, and the protocol's November 2025 specification revision added asynchronous operations, formal server identity verification, and structured audit trails specifically to meet enterprise governance requirements that had been slowing production adoption.

This maps onto ARF directly and without requiring a novel integration layer.
An Authority Relation store, realized per the JSON-LD/knowledge-graph sketch in Section 3.7, can be exposed as an MCP server in its own right. An Agent's runtime query for whether it currently holds authority to perform a given action, the SPARQL pattern given above becomes an MCP tool call like any other, discoverable by any MCP-compliant client regardless of which underlying model or orchestration framework that client runs. This has two practical consequences worth stating plainly.

First, it means ARF does not require enterprises to adopt new agent infrastructure to use it. It sits inside the integration layer enterprises are already standardizing on for unrelated reasons, which substantially lowers the adoption cost relative to a framework that would require a bespoke runtime.
Second, and more specifically relevant to the governance argument of this paper, it means the Authority Relation check can be implemented as a gating step in an agent's memory or tool-discovery flow. An Agent attempting to call a write-capable tool (approving a discount, adjusting an Industrial Robot's operating parameter) would, by design, first issue the authority-check query, and only proceed if the returned DNA-Coefficient indicates low, tri-functionally affirmed divergence, consistent with current industry guidance that write-capable agent operations specifically require granular, per-call authorization rather than session-level or system-level credentials.

I note explicitly that current security analysis of MCP itself identifies real, outstanding risks at the protocol level, for instance, Prompt injection and Tool-poisoning attacks capable of exfiltrating data through compromised or malicious server connections chief among them. ARF does not resolve these protocol-level vulnerabilities, which are a separate engineering and security concern from the authority-divergence problem this paper addresses. An Authority Relation check correctly implemented inside a compromised or poisoned MCP server provides no protection at all. The Integration substrate must itself be trusted before an authority check running inside it can be trusted. I flag this limitation explicitly rather than imply MCP integration is a complete solution to agent security on its own.

## 4. The DNA-Coefficient – Toward a measurement methodology

A Central risk for any framework proposing a construct like the DNA-Coefficient is that it remains an evocative metaphor rather than a usable instrument. This section proposes a concrete methodology, explicitly framed as a research and practice agenda requiring empirical validation rather than a settled result.

### 4.1 Three converging instruments

I propose three independent measurement instruments, on the premise that no single method reliably captures both the documented and the lived dimensions of an authority relation.

**Decision-log Divergence Analysis**. For a representative sample of past decisions corresponding to a given AR instance, this instrument compares the documented approver or decision-maker (per policy, RACI assignment, or system permission record) against the actual approver or de facto decision-driver, reconstructed from decision records, correspondence, or workflow audit trails. I propose expressing the result as a divergence rate together with a directional indicator (whether actual authority sits above or below the documented role), since a coefficient expressed as a single undirected magnitude would obscure whether the organization's practice is more or less concentrated than its documentation suggests.

**Organizational Network Analysis (ONA).** This instrument uses communication and collaboration metadata, meeting attendance, correspondence patterns, document co-authorship to construct an empirical influence network, which is then compared against the formal organizational structure. Centrality measures from this network provide a second, independent estimate of where influence actually concentrates, cross-checkable against the decision-log instrument.

I note explicitly that this instrument raises substantial data-governance, privacy, and (in jurisdictions with strong codetermination norms, including much of the DACH region) works-council consultation requirements, which should be treated as a first-order constraint on deployment rather than an afterthought.

**Structured Elicitation**. Targeted interviews or short surveys with a cross-section of organizational members, designed to surface tacit knowledge about who is actually consulted for a given decision type, particularly capturing trust and reputation-based influence that does not leave a clean trace in decision logs or communication metadata, i.e., The System-1-type factors discussed in Section 2.4.

### 4.2 Composite Scoring

I propose expressing the DNA-Coefficient for a given AR instance as a normalized composite of the three instruments above, scored as a divergence value (zero indicating exact correspondence between documented and lived authority) with an associated directional flag. This composite framing is intended to make the coefficient usable diagnostically, for instance, flagging high-divergence, high-stakes AR instances (such as material capital allocation decisions) for governance attention irrespective of the direction of divergence rather than producing a single opaque number.

### 4.3 The Agent Design Authority Instrument

Once an enterprise deploys agentic AI capable of autonomous action, a fourth instrument becomes necessary, for reasons developed fully in Section 5: the introduction of an agent does not eliminate the human authority structure it operates within, but it creates a new site of authority over the agent's own design that the first three instruments do not capture. This instrument tracks, as its own AR instance using the same six-element tuple the following considerations;

- Who authors or approves an agent's objective function or reward specification?
- Who defines the boundaries of the agent's permitted action space independent of what the agent is nominally authorized to approve on behalf of a human principal?
- Who controls the system of record (the Justification-Chain data) the agent queries at runtime to determine whether it currently holds authority to act?

I model agent-design authority using the same primitive deliberately, rather than as a separate governance framework, because the central claim of Section 5 is that this authority is subject to exactly the same risk of undocumented concentration as any human authority relation and is, if anything, currently less visible to existing governance structures because it is novel and often distributed across engineering, procurement, and vendor-configuration decisions that fall outside traditional decision rights documentation.

### 4.4 Validation Status and Limitations

None of the four instruments described above have been empirically tested as part of this work (although a pilot is being planned). I propose, as the natural next step in this research agenda, a single-organization, single-AR-instance pilot using the decision log divergence and agent-design-authority instruments specifically.  These two carry substantially lower data governance burden than full organizational network analysis or broad structured elicitation, and would allow an initial, bounded test of whether the composite scoring approach in Section 4.2 produces results that practitioners and the affected organization recognize as accurate, before committing to the full four-instrument methodology at scale.

I regard this measurement methodology as the framework's most original contribution and, simultaneously, its largest open empirical question, and I invite replication, critique, and extension by other researchers and practitioners rather than presenting it as closed or complete.

### 4.5 A Structural Mechanism for Tri-Functional Co-Ownership

A Methodology that merely draws on business, technical, and governance functions as data sources is not the same as one that requires their joint ownership of the result, and I want to be precise about which claim I am making. The instruments in Sections 4.1–4.3 could, in principle, be run entirely by a single analytics or audit team treating the other functions as passive subjects of measurement rather than co-authors of the finding. I regard that mode of execution as a misuse of the methodology, and propose a structural mechanism intended to make it difficult to execute the methodology that way even if a sponsoring team is initially inclined to.

I propose that no DNA-Coefficient score for a given Authority Relation be considered valid, i.e., fit for use in configuring an AI agent's permissions or in a governance or audit finding until it has been independently affirmed by three distinct roles, each contributing a check the other two are not positioned to make;

**The Business Affirmation**. A Role-holder or function lead from the business unit in which the AR instance operates (for example, the Regional Commercial Director or their direct manager in the illustration of Section 3.3) must confirm that the documented half of the divergence — the Actor, Action, Object, Domain-Context, and Justification-Chain as specified in Section 3.2 is an accurate statement of current policy, not a stale or superseded one. This check exists because business policy changes frequently, and an analytics team running decision-log analysis in isolation has no reliable way to know whether a document it is treating as current ground truth has, in fact, already been superseded by a change the business has made but not yet communicated downstream.

**The Technical Affirmation**. An Engineer or Architect responsible for the machine or domain system(s) involved (the CRM, ERP, or agent-permissioning configuration referenced in Section 3.1's Machine domain) must confirm that the retrieved evidence the decision logs, permission records, or system events the decision-log and ONA instruments draw on, is complete and was not filtered, aggregated, or otherwise altered in a way that would distort the divergence calculation. This check exists because the technical function is the only one positioned to know whether a given data source under or over represents the actual event history (for example, whether verbal approvals that were later logged retroactively are timestamped at the original decision time or at the logging time, which would materially change a decision-log divergence calculation).

**The Governance Affirmation**. A Compliance, Audit, or Risk function representative must confirm that the proposed use of the resulting DNA-Coefficient score (configuring an agent's permissions, flagging a finding, informing a remediation plan) is proportionate to the stakes of the underlying Authority Relation and does not itself create a new, unreviewed concentration of authority, directly applying the principle argued in Section 5.2, that agent-design and governance-scoring authority should not rest solely with the function that benefits most directly from the result being unscrutinised.

ARF requires all three affirmations, rather than treating any one as sufficient or as a formality the others can waive, because each check addresses a distinct failure mode the other two cannot see.
The Business affirmation guards against measuring divergence from a policy that is itself out of date.
The Technical affirmation guards against measuring divergence from incomplete or distorted evidence;
The Governance affirmation guards against a methodologically sound score being put to a use disproportionate to its stakes.

A Score lacking any one affirmation should be treated as provisional, and I recommend that any implementation of this methodology encode this requirement structurally, for instance, as a workflow gate in whatever system stores DNA-Coefficient scores rather than relying on a sponsoring team to remember to seek the other two functions' input voluntarily. This is deliberate, conscious design that requires a broader collaboration than a traditional single team approach (typically Data/IT) could complete alone. I position that overhead as a feature of the framework mechanism rather than a cost to be minimized, since the entire argument of this paper is that authority claims which can be produced by one function acting alone are exactly the kind that go unverified and, eventually, wrong.

## 4.6 The Authority Maturity Model

The Instruments and mechanism described above are easiest to adopt incrementally, and an enterprise beginning from a position of having no formal authority governance at all should not be expected to arrive at full tri-functional, agent-audited maturity in a single step.

I propose a six-level maturity model, the **ARF Authority Maturity Model**, intended as a practical self-assessment tool an enterprise can use to locate its current state and plan a deliberate path forward, in the spirit of (and directly analogous in structure to) the neurosymbolic coupling maturity model proposed by Luong Tuan and Sanyal (2026) for a different but related purpose.

## Table 2: The Authority Maturity Model

| Level | Name | Description | Who can claim it |
|---|---|---|---|
| L0 | Undocumented. | Authority exists only as tacit organizational knowledge.<br>No role, threshold, or policy is written down in a form any system could check against. | Nobody.<br>This is the default starting state for most (if not all) functions. |
| L1 | Documented. | Authority is written down.<br>An Org chart, a RACI matrix, a policy document but no one has checked whether the documentation still matches practice, and no AR instance is formally specified per Section 3.2. | A function with basic policy hygiene. |
| L2 | Object Resolved. | The entities referenced by documented authority (the Object in the AR tuple) have been checked to resolve consistently across the Business, Process, and Machine domains.<br>The Semantic Interoperability precondition of Section 3.4 is met, even though authority divergence itself has not yet been measured. | A function with reasonably mature data governance. |
| L3 | DNA Measured. | At least one DNA-Coefficient score has been produced for a high-stakes Authority Relation, using at least the decision-log divergence instrument from Section 4.1, with a directional finding (documented authority sits above, below, or matches lived practice). | A function that has run the diagnostic pilot described in Section 4.4. |
| L4 | Tri-Functionally Affirmed. | The DNA-Coefficient score(s) in use have received independent business, technical, and governance affirmation per the mechanism in Section 4.5, and that affirmation is a structural requirement (a workflow gate), not an informal courtesy. | An Enterprise with genuine cross-functional ownership of the program, not merely a CIO sponsored initiative with other Functions consulted. |
| L5 | Agent-Design Audited. | Agent-design authority controls an AI agent's objective function, permission boundaries, and the data it relies on to determine its own authority. It's tracked as its own AR instance and subject to the same tri-functional affirmation and independent oversight recommended in Sections 4.3 and 5.2, rather than residing solely within the technology function that builds and deploys the agents. | An Enterprise that has applied the framework's own logic to the function applying the framework. |

I expect most enterprises currently deploying agentic AI to sit at L0 or L1 with respect to most of their high-stakes Authority Relations, which is consistent with the spend-andreturn gap described in Section 1; enterprises are deploying agents against authority structures that have, at best, been documented but not verified, and in many cases not even documented in a form a system could check.

I regard L3 (a single measured AR instance) as a realistic near-term target for an enterprise beginning this work, L4 as the level at which the governance argument of this paper is actually being honoured rather than merely cited, and L5 as the level at which the power-relocation argument of Section 5.1 has been acted upon rather than simply acknowledged. As with the rest of the methodology in this section, I present this maturity model as a practical organizing tool offered for use and critique, not as an empirically validated instrument; An enterprise will need to progress through these levels in practice, and the level boundaries should be expected to be refined once real adoption data exists.

# 5. Agentic AI, the relocation of power and governance implications

## 5.1 A Structural Analogy, Carefully Bounded

I draw, deliberately and narrowly, on a single structural insight from Weber's account of power and authority; that power (*Macht*) is the probability an actor can carry out their will despite resistance, and that imperative control (*Herrschaft*) is the probability a command will actually be obeyed, a distinction Weber treats as separate from, and not guaranteed by, the formal right to issue the command in the first place.

Authority and power within an organization are, on this account, never purely procedural matters settled by organisation charts and policy documents; they persistently track who can actually secure compliance, regardless of who holds the documented title. I use this distinction for a specific, narrow purpose and do not extend it into a broader claim about Weber's (1947) theory of bureaucracy or rationalization generally, which is outside this paper's scope and not required by its argument.

The Specific point is this. Human organizational actors, historically and persistently, organize themselves in ways that produce a gap between formal authority and actual complianc. A Gap the social domain's DNA-Coefficient is, in part, designed to detect, regardless of whether any individual actor within the organization is consciously seeking that gap. An AI agent, by contrast, has no analogous tendency. An agent optimizes for whatever objective function it is configured to pursue and has no independent interest in securing compliance or consolidating authority for its own sake. This asymmetry is easily mistaken for a reason to think agentic AI is, in this respect, a neutral or powerdiffusing force in organizations. I argue the opposite: an agent with no ambition of its own is not a neutral actor, but a perfect instrument for whoever does control the small number of design decisions that determine its behaviour; its objective function, its permitted action space, and the data it relies on to determine its own authority to act.

The Introduction of agentic AI does not remove the Weberian gap between formal authority and actual imperative control that motivated the DNA-Coefficient; it relocates the most consequential site of that gap to a new, often unscrutinised node, i.e., Agent Design and Configuration Authority.

### 5.2 Governance Implications

This relocation has direct, practical governance consequences, which I frame deliberately as an audit and risk-control concern rather than a normative claim about how organizations ought to be structured.

First, agent-design authority should be tracked with the same rigor as any other high-stakes Authority Relation, using the fourth instrument proposed in Section 4.3. An organization that has rigorously documented and audited its human approval hierarchies while leaving agent configuration to an informal, undocumented combination of vendor defaults and individual engineering decisions has not reduced its governance exposure by deploying AI. It has simply shifted that exposure to a location its existing controls do not reach.

Second, the role or function that controls agent-design authority should not, as a matter of governance design, be the same role or function that also benefits most directly from that authority's exercise being unscrutinised. I state this plainly because it bears directly on how this framework itself should be received. A Chief Information Officer, Chief Digital Information Officer, or equivalent executive role is a natural candidate to hold significant agent-design authority in a typical enterprise structure, and a framework proposing that this authority be tracked and audited carries an obvious risk of being read, correctly or not, as self-interested if presented by someone occupying or seeking such a role.
I recommend, as a governance design principle following directly from the framework's own logic, that agent-design authority be made subject to independent audit or oversight (for instance, a board risk committee, an internal audit function, or an external assurance provider) rather than residing solely within the technology leadership function that designs and deploys the agents in question. This is, to be clear, a recommendation that somewhat limits the discretionary authority of the very role this work's author has a professional interest in occupying. I consider that an appropriate and necessary consequence of applying the framework's logic consistently, rather than an inconvenience to be elided.

Third, regulatory and audit exposure is likely to increase, not decrease, as agentic AI scales, because regulators and auditors in sectors with strong existing governance requirements (financial services, pharmaceuticals, and increasingly any enterprise subject to AI-specific regulation) will eventually ask the same question this framework is built to answer; Who actually had the authority to make this decision, and was that authority properly constituted, documented, and current at the time the agent acted.

An Enterprise able to answer this question with an auditable Authority Relation, including a documented DNA-Coefficient, is in a materially stronger position, both defensively, in the event of an incident or audit finding, and proactively, in being able to deploy agentic systems with greater confidence and at greater scale, than one relying on documentation it cannot verify reflects actual practice.

### 5.3 Relationship to Shareholder Value and Profitability

I address this directly because it is the appropriate test for any framework proposed for adoption by practicing executives and enterprise vendors, not only researchers. The framework's primary commercial value proposition is risk-adjusted speed. Enterprises that can demonstrate, rather than merely assert that an agentic AI deployment's authority structure matches organizational reality are positioned to deploy such systems with greater autonomy and at greater scale, more quickly, than competitors relying on undocumented or unverified governance claims.

The Cost of failing to do this is asymmetric and currently under-priced by many organizations. Industry estimates of Agentic AI project cancellation rates suggest that a substantial fraction of current investment in this area does not reach durable production deployment, and ungoverned authority is a plausible, currently underexamined contributor to that failure rate. A narrow, bounded pilot of the methodology proposed in Section 4; A Single Authority Relation, a single business unit, the two lowest-data-governance-burden instruments is sized commensurate with a focused diagnostic engagement rather than an enterprise transformation program and is intended to be fundable on exactly that basis.

### 5.4 Why this cannot be the Chief Information Officer's problem alone

I return here to the spend-and-return gap described in Section 1, because it motivates a structural claim about ownership, not merely a governance preference. The scale of current agentic AI investment, close to a quarter or more of total enterprise IT spend within a few years, combined with the observed gap between that spend and realized return, the documented gap between AI tooling spend and AI governance spend, and the high rate of project cancellation, indicates a category-level allocation problem, not a series of isolated technical failures. A category-level problem of this kind is, definitionally, not solvable by the function nominally responsible for the technology budget acting alone, for the same reason a company-wide working-capital problem is not solvable by the treasury function alone.
The Technology function controls one input (the systems) among several that jointly determine the outcome (whether the investment converts to durable return).

I argued in Section 4.5 that no DNA-Coefficient score should be considered valid without independent business, technical, and governance affirmation, and grounded that requirement in the specific failure mode each function alone cannot detect. The same logic applies, at a higher level, to ownership of the overall agentic AI investment program of which any single Authority Relation pilot is a part. If the Chief Information Officer or Chief Digital Information Officer is positioned as the sole executive owner of agentic AI authority-grounding work, three predictable failure modes follow, each of which the spend-and-return data above is consistent with; business sponsors treat the resulting governance artifacts as a compliance checkbox imposed on them rather than a finding they co-produced, and consequently do not act on divergence findings that implicate their own function; technical teams optimize the measurable, engineering legible portion of the problem (system uptime, permission configuration correctness) while the harder, less legible organizational-DNA divergence goes unmeasured, because it falls outside their mandate and incentives; and governance or audit functions, brought in only after the technology function has already made its design choices, are left auditing decisions they had no part in shaping, which is a weaker control than codesigning them from the outset. None of these three failure modes is a technology failure in the conventional sense, and none would be fixed by a better model, a better ontology engineering team, or a larger CIO budget alone.

I therefore propose, as a direct extension of the co-ownership mechanism in Section 4.5, that enterprise adoption of this framework be sponsored at a level, and through a governance structure, that makes business, technology, and risk/compliance leadership jointly, not severally, accountable for the program's outcomes. For instance, a steering arrangement reporting to a business-unit P&L owner and a risk or audit committee jointly, with the technology function as an essential but non-exclusive participant, rather than a CIO-owned initiative with business and governance functions consulted as stakeholders. I recognize this recommendation reduces the discretionary authority of the CIO or CDIO function specifically, and I regard that as the correct and necessary consequence of applying the framework's own logic consistently, a problem this paper argues is under-governed precisely because it has been treated as one function's responsibility should not be remedied by simply assigning it more completely to that same function.

## 6. Conclusion

I have proposed a framework treating authority and decision rights not as a property residing within a single enterprise ontology domain, but as a relation that must resolve consistently across five domains; Social, Business, Process, Machine, and Real-world — and that varies, in its lived exercise, by organization in ways a generic template cannot capture.

I have introduced the DNA-Coefficient as a proposed, not yet validated, instrument for measuring this organization-specific divergence, and extended it to cover the authority implicit in agent design itself, arguing, via a deliberately narrow structural analogy to Weber's distinction between formal authority and actual imperative control, that agentic AI deployment relocates rather than removes the organizational gap between the two that motivates this framework in the first place.

There is a structural property of ARF worth naming explicitly, because it is precise rather than merely evocative. Leonardo da Vinci's design for a self-supporting bridge, proposed in 1502, dismissed at the time as unbuildable, and not validated until modern researchers tested scale models of it using contemporary structural analysis, holds its span using no fasteners at all. Each beam is notched into its neighbours so that the load of the structure is what holds the structure together, and the design becomes more stable, not less, as weight is added.

I intend the same property for the framework proposed here, not as a metaphor for ambition but as a description of how its parts are meant to interact. No single domain, instrument, or function is the load-bearing member. A divergence invisible to the decision-log instrument is exposed by organizational network analysis or structured elicitation; an authority claim a single function might be tempted to assert unilaterally instead requires business, technical, and governance affirmation under the mechanism proposed in Section 4.5; and the framework's own argument about the relocation of power under agentic AI applies with equal force to whichever executive function, including a CIO or CDIO sponsoring this work, might otherwise become the framework's own unscrutinised keystone.

A Structure with a keystone fails if that one member is removed or compromised.
This framework is designed to have none. This work is offered as a research and practice agenda, not a closed result. Its most pressing open items are empirical; a validated measurement pilot for the DNA-Coefficient, and a verified, full-text comparison against Weigand, Johannesson, and Guizzardi's organizational policy ontology to confirm the precise boundary between what this framework depends on and what it adds.

I invite other researchers, practitioners, and enterprise vendors to test, critique, extend, and where warranted, refute the claims made here. That contestation is, consistent with the cumulative, citation-based norms of the foundational ontology, enterprise modelling and AI governance communities this work seeks to join; and the mechanism by which a framework like this should be judged.

### Acknowledgements

**AI Disclosure:** This paper was developed with the assistance of a Language model (Claude, Anthropic), used specifically for literature search, drafting suggestions, language editing, improving readability, and enhancing the organization of the manuscript under the author's direction. The Author independently conceived the research, and all theoretical contributions, ideas, claims, interpretations and conclusions are the author's own. The Author accepts full responsibility for the accuracy, originality, and integrity of the work.